\documentclass{article}

\usepackage{arxiv}

\usepackage[utf8]{inputenc} 
\usepackage[T1]{fontenc}    
\usepackage{hyperref}       
\usepackage{url}            
\usepackage{booktabs}       
\usepackage{amsfonts}       
\usepackage{nicefrac}       
\usepackage{microtype}      
\usepackage{lipsum}		
\usepackage{graphicx}
\usepackage{natbib}
\usepackage{doi}
\usepackage{multirow}

\title{LongQLoRA: Efficient and Effective Method to Extend Context Length of Large Language Models}

\author{ 
	Jianxin Yang\\
    \texttt{yangjx57@mail2.sysu.edu.cn}
}

\hypersetup{
pdftitle={LongQLoRA: Efficient and Effective Method to Extend Context Length of Large Language Models},
pdfsubject={},
pdfauthor={Jianxin Yang},
pdfkeywords={},
}

\begin{document}
\maketitle

\begin{abstract}
We present LongQLoRA, an efficient and effective method to extend context length of large language models with less training resources. LongQLoRA combines the advantages of Position Interpolation, QLoRA and Shift Short Attention of LongLoRA. With a single 32GB V100 GPU, LongQLoRA can extend the context length of LLaMA2 7B and 13B from 4096 to 8192 and even to 12k within 1000 finetuning steps. LongQLoRA achieves competitive perplexity performance on PG19 and Proof-pile datasets, our model outperforms LongLoRA and is very close to MPT-7B-8K within the evaluation context length of 8192. We collect and build 39k long instruction data to extend context length of Vicuna-13B from 4096 to 8192 and achieve good performance both in long and short context generation task. We also do some ablation experiments to study the effect of LoRA rank, finetuning steps and attention patterns in inference.The model weights, training data and code are avaliable at \url{https://github.com/yangjianxin1/LongQLoRA}.

\end{abstract}


\section{Introduction}
With the advent of LLaMA\citep{touvron2302llama}, a lot of open-source works based on LLaMA have emerged, such as Alpaca\citep{taori2023stanford}, Vicuna\citep{chiang2023vicuna}, Gunaco\citep{dettmers2023qlora}, WizardLM\citep{xu2023wizardlm}, etc. Through instruction tuning and RLHF, these models have achieved excellent performance on many NLP tasks, and even surpass ChatGPT evolved from InstructGPT\citep{ouyang2022training} on some specific tasks. The LLaMA-series models are trained on a pre-defined context length, such as 2048 of LLaMA and 4096 of LLaMA2\citep{touvron2023llama}. The positional encoding of LLaMA-series models is RoPE\citep{su2021roformer}, which has weak extrapolation properties. When the input length of LLaMA exceeds the pre-defined context length, the perplexity of the model increases sharply and its performance also drops obviously. For some tasks with long input context, the pre-defined short context length limits model's performance, such as multi-document QA, book summarization, long dialogue summarization, etc.

To increase the context length of LLaMA2\citep{touvron2023llama}, the most straightforward way is to further pretrain LLaMA2 with longer text just like MPT-7B-8K\citep{team2023introducing}. However, this method requires a lot of training GPUs and is slow to converge. In order to solve this problem, Meta proposes Position Interpolation(PI)\citep{chen2023extending}, which uses 32 A100 GPUs to extend the context length of LLaMA from 2048 to 8192. It only finetunes LLaMA with 1000 steps and achieves good performance. Focused Transformer(FOT)\citep{tworkowski2023focused} presents LongLLaMA with 256k context length trained on 128 TPUs, FOT is a plug-and-play extension method, and the model can be easily extrapolated to longer sequences. For example, a model trained on 8k context length can be easily extrapolated to 256k. LongLoRA\citep{chen2023longlora} proposes shift short attention, combines Position Interpolation and LoRA\citep{hu2021lora} to implement a more efficient method. It extends LLaMA2's context length from 4096 to 100k on 8 A100 GPUs.

\begin{figure}
	\centering
    \includegraphics[scale=0.5]{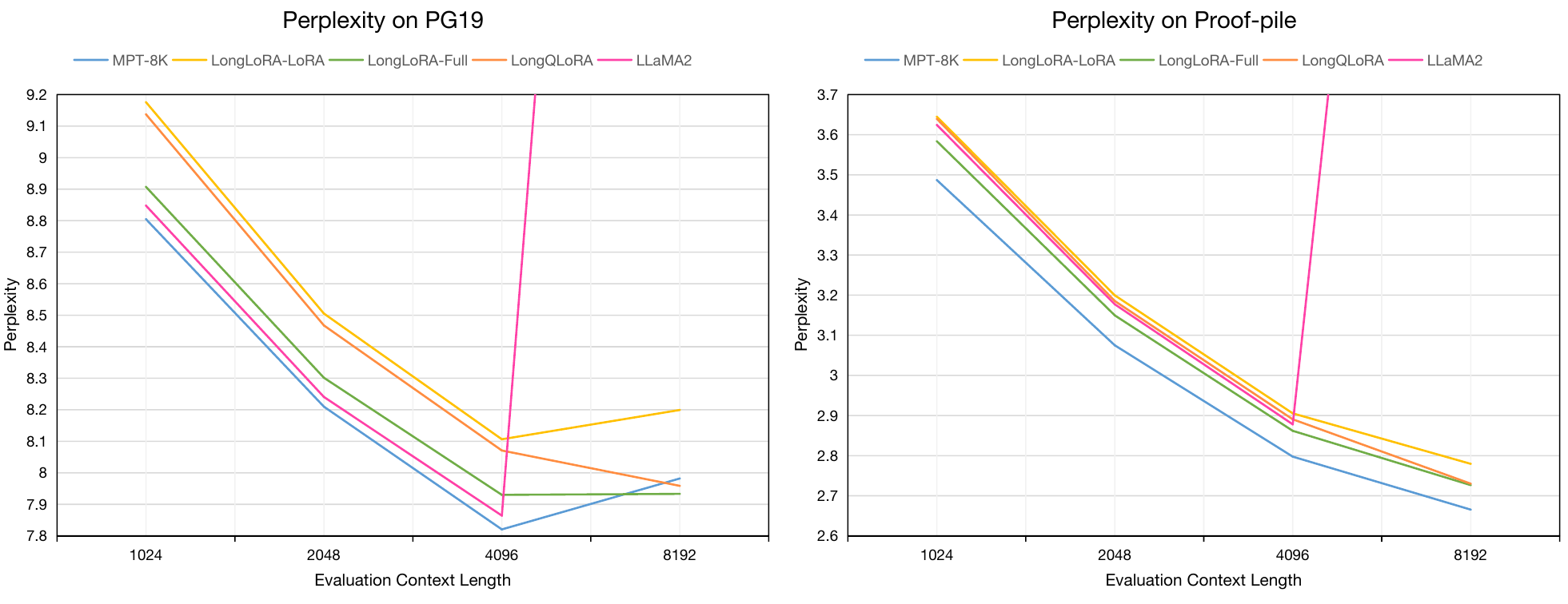}
	\caption{Evaluation perplexity of 7B models on PG19 validation and Proof-pile test datasets in evaluation context length from 1024 to 8192. All models are quantized to 4-bit in inference. LongQLoRA is finetuned based on LLaMA2-7B for 1000 steps with RedPajama dataset on a single V100 GPU. ‘LongLoRA-Full’ and ‘LongLoRA-LoRA’ mean LLaMA2-7B published by LongLoRA with full finetuning and LoRA finetuning respectively. MPT-7B-8K are better than LLaMA2, LongLoRA and LongQLoRA in context length from 1024 to 4096. LLaMA2-7B has very poor performance beyond the pre-defined context length of 8192. LongQLoRA outperforms LongLoRA-LoRA on both datasets in context length from 1024 to 8192. In context length of 8192, LongQLoRA is extremely close to LongLoRA-Full on Proof-pile test dataset, even better than MPT-7B-8K on PG19 validation dataset.
}
	\label{fig:perplexity}
\end{figure}

However, both Position Interpolation and FOT require much computation resources, 32 A100 GPUs and128 TPUs respectively. Although LongLoRA can save a lot of training resources, it still spends 8 A100 GPUs. It's unaffordable for common researchers. Is there any other method can further save training resources and ensure the performance? We believe that QLoRA\citep{dettmers2023qlora} is a good option. QLoRA is an efficient finetuning method that quantizes pretrained large language model to 4-bit to reduce model's memory footprint, and then adds a small set of learnable Low-rank adapter weights. This method can be used to finetune LLaMA 65B on a single 48GB GPU and preserve full 16 bit finetuning task performance.

\begin{table}
	\caption{Evaluation perplexity on PG19 validation and Proof-pile test datasets in evaluation context length of 8192. All models are quantized to 4-bit in inference. ‘LongLoRA-Full’ and ‘LongLoRA-LoRA’ mean LLaMA2-7B published by LongLoRA with full finetuning and LoRA finetuning respectively. LLaMA2-7B has very poor performance beyond the pre-defined context length. LongQLoRA outperforms LongLoRA-LoRA and close to LongLoRA-Full and MPT-7B-8K.
}
	\centering
	\begin{tabular}{ccc}
		\toprule
        \multirow{2}{*}{Model} & \multicolumn{2}{c}{Dataset} \\
        		~ & PG19 & Proof-pile \\
		\midrule
        LLaMA2-7B & >$10^3$  & >$10^3$     \\ 
		MPT-7B-8K & 7.98  & 2.67     \\ 
		LongLoRA-LoRA-7B-8K  & 8.20 & 2.78    \\
		LongLoRA-Full-7B-8K  & 7.93  & 2.73   \\
        LongQLoRA-7B-8K  & 7.96 & 2.73      \\
		\bottomrule
	\end{tabular}
	\label{tab:summary-result}
\end{table}

In this work, we present LongQLoRA, a memory-efficient and effective method to extend context length of LLaMA-series models. With LongQLoRA, we can extend context length of LLaMA2 from 4096 to 8192, even to 12k on a single V100 with 32GB memory. LongQLoRA combines the advantages of Position Interpolation, QLoRA and shift short attention of LongLoRA. There are some differences between LongQLoRA and LongLoRA in details. In LongQLoRA, we set LoRA rank as 64, Low-rank adapter weights are added to all layers and we don't train word embeddings and normalization layers\citep{ba2016layer}. We extend the context length of LLaMA2\citep{touvron2023llama} 7B and 13B to 8192 on a single V100 with 32GB memory. 

As shown in Table \ref{fig:perplexity}, LLaMA2-7B has very poor performance beyond the pre-defined context length and the perplexity of LongQLoRA drop significantly compared with LLaMA2-7B. LongQLoRA has better performance than LongLoRA on PG19\citep{rae2019compressive} validation and Proof-pile\citep{rae2019compressive} test datasets, and is very close to MPT-7B-8K. Note that MPT-7B-8K starts from MPT-7B\citep{team2023introducing}, updates the sequence length to 8k and train for an additional 500B tokens, results in a total of 1.5T tokens of text and code, which cost a lot of training resources and data. Figure \ref{fig:perplexity} shows that LongQLoRA also outperforms LongLoRA in evaluation context length from 1024 to 8192 both on PG19 and Proof-pile dataset.

In addition, we build a long instruction dataset with 39k data, mainly including book summarization, Natural Questions\citep{kwiatkowski2019natural}, subset of LongQA\citep{chen2023longlora} and Evol-Instruct of WizardLM\citep{xu2023wizardlm}. We extend the context length of Vicuna-13B-V1.5\citep{chiang2023vicuna} from 4096 to 8192 with the long instruction dataset and achieve good performance.

In summary, the contributions of our work are as follows:
\begin{itemize}
    \item We present LongQLoRA, a memory-efficient and effective method to extend context length of LLaMA2, which combines QLoRA, Position Interpolation and Shift Short Attention.
    \item We evaluate the performance of LongQLoRA on PG19 and Proof-pile datasets and demonstrate the effectiveness of our method. As for LLaMA2 7B, LongQLoRA outperforms LongLoRA and is close to MPT-7B-8K.
    \item We collect and build 54k long pretraining text and 39k long instruction data. We make our data, code and model weights public avaliable for further research.
\end{itemize}

\section{Method}
\label{sec:headings}

\begin{figure}
	\centering
    \includegraphics[scale=0.45]{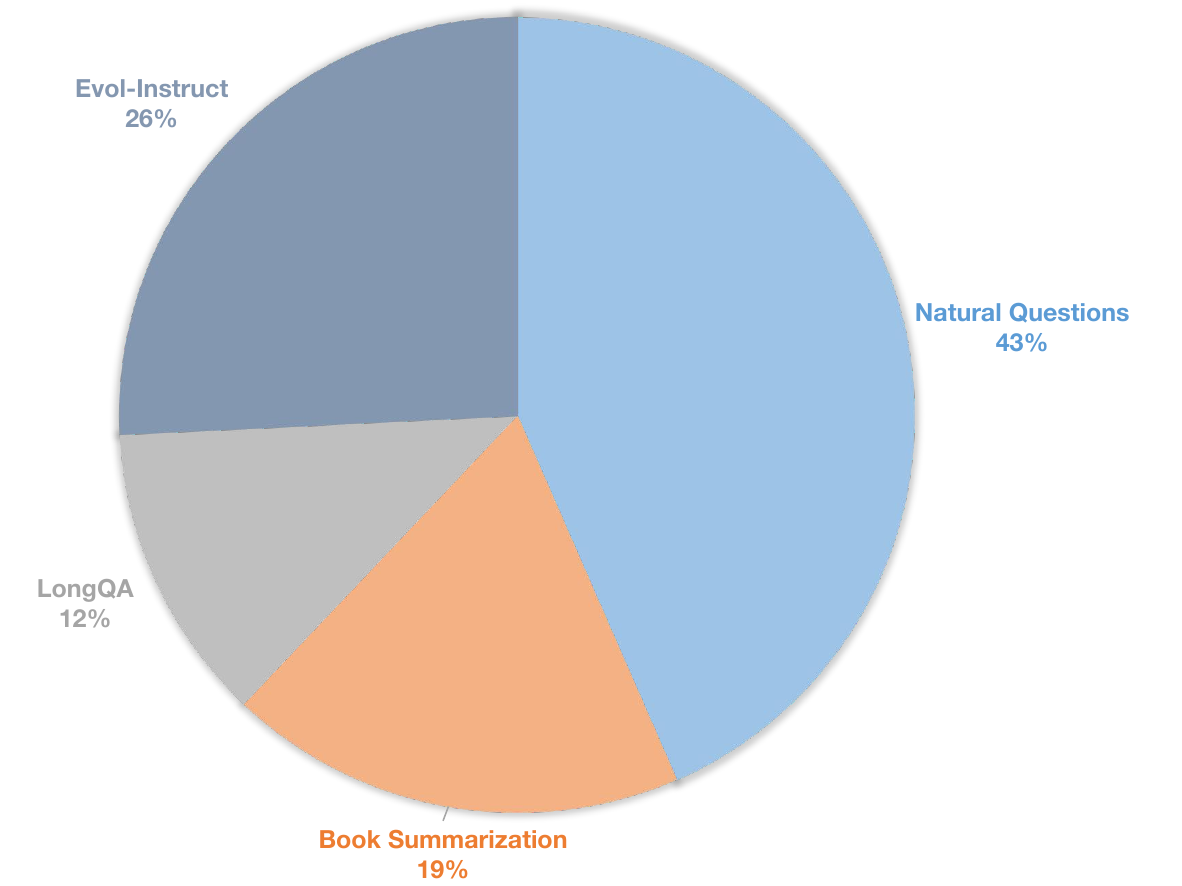}
	\caption{Distribution of long instruction tuning dataset. The dataset contains 39k instruction data, mainly including book summarization, Natural Questions, subset of LongQA and Evol-Instruct of WizardLM. In order to adapt to the target length of 8192, the max sequence length of the dataset is 8192. We add some short data sampled from Evol-Instruct to avoid the degration at short instruction following.
}
	\label{fig:data-distribution}
\end{figure}

\begin{table}
	\caption{Evaluation perplexity of models on PG19 validation dataset in different evaluation context length. All models are quantized to 4-bit only in evaluation context length of 8192, otherwise 16-bit is used.
}
	\centering
	\begin{tabular}{ccccc}
		\toprule
        \multirow{2}{*}{Model} & \multicolumn{4}{c}{Evaluation Context Length} \\
        		~ & 1024 & 2048 & 4096 & 8192 \\
		\midrule
        LLaMA2-7B & 8.85  & 8.24 & 7.86 & >$10^3$    \\ 
		MPT-7B-8K & 8.80  & 8.21 & 7.82 & 7.98     \\ 
		LongLoRA-LoRA-7B-8K  & 9.18 & 8.51   & 8.11 & 8.20   \\
		LongLoRA-Full-7B-8K  & 8.91  & 8.30 & 7.93 & 7.93  \\
        LongQLoRA-7B-8K  & 9.14 & 8.47   & 8.07 & 7.96   \\
		\bottomrule
	\end{tabular}
	\label{tab:pg19}
\end{table}

\begin{table}
	\caption{Evaluation perplexity of models on Proof-pile test dataset in different evaluation context length.
}
	\centering
	\begin{tabular}{ccccc}
		\toprule
        \multirow{2}{*}{Model} & \multicolumn{4}{c}{Evaluation Context Length} \\
        		~ & 1024 & 2048 & 4096 & 8192 \\
		\midrule
        LLaMA2-7B & 3.62  & 3.18 & 2.88 & >$10^3$    \\ 
		MPT-7B-8K & 3.49  & 3.08 & 2.80 & 2.67     \\ 
		LongLoRA-LoRA-7B-8K  & 3.64 & 3.20   & 2.91 & 2.78   \\
		LongLoRA-Full-7B-8K  & 3.58  & 3.15 & 2.86 & 2.73  \\
        LongQLoRA-7B-8K  & 3.64 & 3.19   & 2.89 & 2.73   \\
		\bottomrule
	\end{tabular}
	\label{tab:proof-pile}
\end{table}

\subsection{Background}


\paragraph{Position Interpolation.}
Meta proposes Position Interpolation\citep{chen2023extending} to extend the context length of RoPE-based\citep{su2021roformer} large language models such as LLaMA. With Position Interpolation, LLaMA's context window sizes can be extended from 2048 to 32768 after finetuning within 1000 steps. When extending the context length, Position Interpolation doesn't directly train beyond the pre-defined context length. Instead, it aligns the target max position index with the pre-defined max position index, and then performs positional encoding interpolation in the pre-defined positional space.

With Position Interpolation and only 1000 steps of finetuning, LLaMA can be easily extended to long target length and achieves good perplexity in PG19 dataset. 

\paragraph{Shift Short Attention.}
Different from standar self-attention\citep{chen2023longlora}, shift short attention is a sparse local attention mechanism proposed by LongLoRA which is effective and efficient in finetuning. Standar self-sttention\citep{vaswani2017attention} has $O(n^2)$ computational complexity which costs high GPU memory. Instead, shift short attention splits input tokens into groups and only computes the attention in each group individually. In order to enhance the information interaction between adjacent groups, it also computes the attention between the neighbouring groups. With the sparse local attention mechanism, shift short attention can save much GPU memory. Assuming the input tokens are split into $g$ groups, the computational complexity can be reduced from $O(n^2)$ to $O((n/g)^2)$.

\paragraph{QLoRA.}
QLoRA\citep{dettmers2023qlora} is an efficient finetuning method for large language model proposed by the University of Washington. QLoRA can finetune LLaMA\citep{touvron2302llama} 65B on a single 48GB GPU and preserve full 16-bit finetuning task performance. Guanaco\citep{dettmers2023qlora} finetuned with QLoRA can reach 99.3\% of the performance level of ChatGPT.

QLoRA quantizes the weights of pretrained model to 4-bit, and then adds learnable Low-rank\citep{hu2021lora} Adapter weights. QLoRA freezes the pretrained model and only finetunes the LoRA adapters. The main contributions of QLoRA include 4-bit NormalFloat, Double Quantization and Paged Optimizers. 4-bit NormalFloa is a theoretically optimal 4-bit quantization data type, which is better than FP4 and Int4. Double Quantization can save more GPU memory than the previous model quantization method. It can save an average of 0.37 bits per parameter, for LLaMA 65B, it can save approximately 3GB GPU memory. Paged Optimizers uses NVIDIA unified memory to avoid gradient checkpointing memory spikes when processing long sequences of mini-batches.

Based on the above optimization, QLoRA can finetune large language models with very limited GPU resources, and achieve excellent performance close to full finetuning.

\subsection{LongQLoRA}
LongQLoRA combines the advantages of Position Interpolation\citep{chen2023extending}, QLoRA\citep{dettmers2023qlora} and Shift Short Attention of LongLoRA\citep{chen2023longlora}. Firstly, we use Position Interpolation to extend the context length of LLaMA2\citep{touvron2023llama} from 4096 to the target size. In order to save more GPU memory, during finetuning, we use QLoRA to quantize the weights of base model to 4-bit. To further save GPU memory, we also use Shift Short Attention in finetuning with group size 1/4 of the target context length.

In order to recover the performance lost due to imprecise quantization, we add LoRA\citep{hu2021lora} adapters on all layers, and the LoRA rank is 64. Different from LongLoRA, LongQLoRA can achieve better performance even without training word embeddings and normalization layers\citep{ba2016layer}. This is due to the fact that we add more LoRA adapters and use larger LoRA rank.

We find that it achieves better inference performance with standar global attention\citep{vaswani2017attention} compared with Shift Short Attention, so we uniformly use standar global attention in inference. It means models finetuned with LongQLoRA can be perfectly compatible with existing technologies in inference, such as Flash Attention\citep{dao2022flashattention}, vLLM\citep{kwon2023efficient} and so on. This is also one of the advantage of Shift Short Attention, we don’t cost additional adaptation effort in inference.

\section{Experiment}
\label{sec:experiment}

\begin{table}
	\caption{Ablation on LoRA rank settings in LongQLoRA. All the models are 7B size. The target context length, evaluation context length and sliding window size are 8192. With the LoRA rank increases, the perplexity of LongQLoRA drops gradually. When LoRA rank is set to 64, LongQLoRA outperforms LongLoRA-LoRA, MPT-7B-8K and close to LongLoRA-Full. 
}
	\centering
	\begin{tabular}{cccccccc}
		\toprule
		\multirow{2}{*}{Model} & \multicolumn{4}{c}{LoRA Rank} & \multirow{2}{*}{LongLoRA-LoRA} & \multirow{2}{*}{LongLoRA-Full} & \multirow{2}{*}{MPT-7B-8K}  \\
            ~& 8 & 16 & 32 & 64 & & \\
		\midrule
		PPL & 8.01  & 7.97 & 7.96 & 7.96 & 8.20 &  7.93 & 7.98  \\
		\bottomrule
	\end{tabular}
	\label{tab:lora rank}
\end{table}

\begin{table}
	\caption{Ablation on finetuning steps of LongQLoRA. Models are finetuned based on LLaMA2-7B with RedPajama dataset. The results are evaluated in perplexity on PG19 validation dataset. Only after 100 finetuning steps, the perplexity drops dramatically. As the number of finetuning steps increases, the perplexity further decreases and converges.}
	\centering
	\begin{tabular}{cccccccccccc}
		\toprule
		\multirow{2}{*}{Model} & \multicolumn{11}{c}{Number of finetuning steps}\\
            ~ & 0 & 100 & 200 & 300 & 400 & 500 & 600 & 700 & 800 & 900 & 1000 \\
		\midrule
		LongQLoRA-7B-8K & 9.45  & 8.14 & 8.05 & 8.00 & 8.01 & 8.00 & 7.98 & 7.99 & 7.99 & 7.97 & 7.96    \\
		\bottomrule
	\end{tabular}
	\label{tab:finetuning steps}
\end{table}
We mainly conduct experiments on 7B and 13B models, and only use a single V100 GPU with 32GB memory during the whole experiment. We extend the context length of LLaMA2-7B and Vicuna-13B from 4096 to 8192.

\subsection{Experiment Settings}
\paragraph{Training Procedure.}
We firstly use Position Interpolation technology to extend the context length from 4096 to 8192. As for QLoRA, we quantize the weights of base model to 4-bit NormalFloat\citep{dettmers2023qlora}, LoRA rank is set as 64 and add LoRA adapters to all layers, including q\_proj, k\_proj, v\_proj, up\_proj, down\_proj, gate\_proj and o\_proj. Finally, the number of trainable parameters is about 1.5 million and 2.5 million for 7B and 13B model respectively. We use Paged Optimizers\citep{dettmers2023qlora}, the learning rate is set to 2e-4 and 1e-4 for 7B and 13B model respectively, we use constant learning rate with warmup\citep{fradkin2010effects}, warmup step is 20. We set per-device batch size as 1 and gradient accumulation steps as 16, which means the global batch size is 16 with only a single GPU. We use Deepspeed Zero2\citep{rasley2020deepspeed} strategy during finetuning. We finetune 1000 steps for LLaMA2-7B while 1700 steps for Vicuna-13B.

When finetuning LLaMA2-7B, we use the next token prediction task, and only compute the cross entropy loss on the target part when finetuning Vicuna-13B.

For shift short attention, we set the group size to 1/4 of model's max context length, we use shift short attention during finetuning and standard global attention during inference.

\paragraph{Datasets.}
We sample about 54k long text from Redpajama\citep{together2023redpajama} dataset for finetuning pretrained models, whose token lengths ranging from 4096 to 32768.

We conduct perplexity evaluation on PG19\citep{rae2019compressive} validation dataset and Proof-pile\citep{rae2019compressive} test dataset for pretrained model. We use standard global attention\citep{vaswani2017attention} and the sliding window is the same as evaluation context length during evaluation. When the evaluation length is greater than or equal to 8192, we quantize the model weights to 4-bit, otherwise float16 is used in inference. 

In addition, we also build a long context instruction dataset for supervised finetuning chat models. This dataset contains 39k instruction data, mainly including book summarization, Natural Questions\citep{kwiatkowski2019natural}, subset of LongQA\citep{chen2023longlora} and Evol-Instruct of WizardLM\citep{xu2023wizardlm}. In order to adapt to the target context length of 8192, the max token number of each data is 8192. The distribution of dataset is shown as Figure \ref{fig:data-distribution}.

\begin{figure}
	\centering
    \includegraphics[scale=0.51]{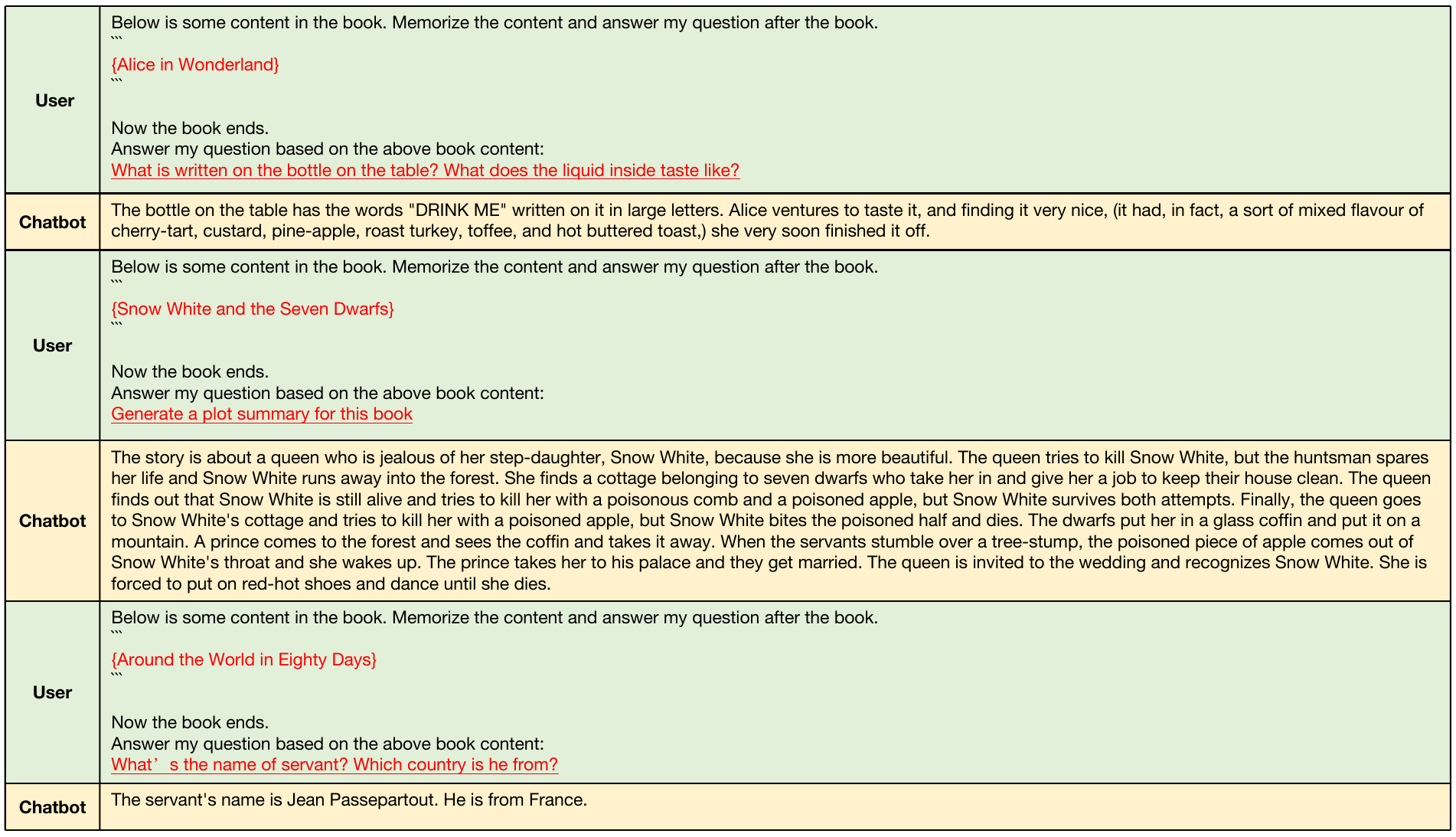}
	\caption{Examples of long context generartion. The input context lengths are between 4096 and 8192 which are larger than original context length of LLaMA2.
}
	\label{fig:longdemo}
\end{figure}

\begin{figure}
	\centering
    \includegraphics[scale=0.51]{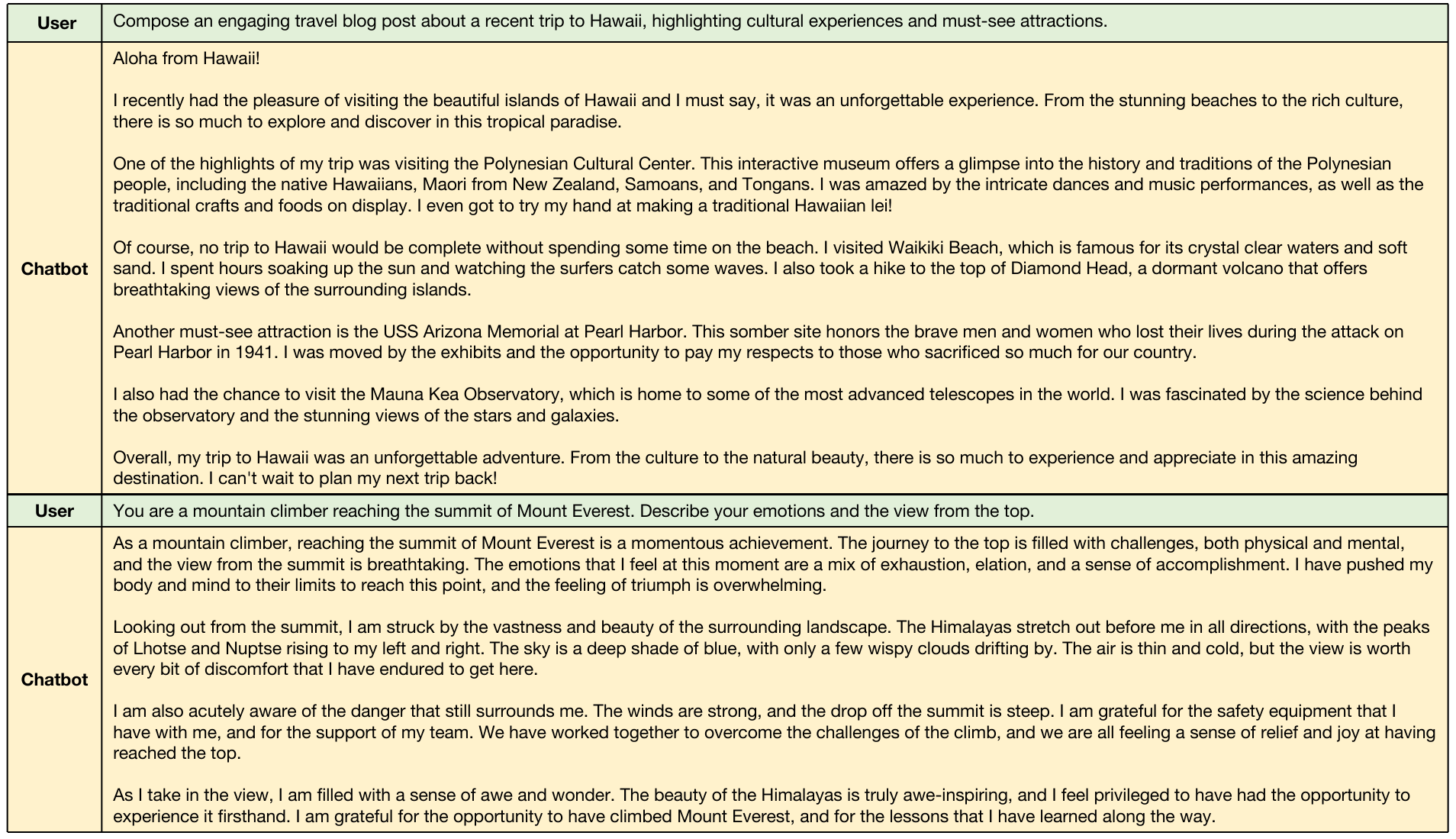}
	\caption{Examples of short context generartion. Our model also keep the good performance in short context generation which means that LongQLoRA doesn't harm the performance of short instruction following.
}
	\label{fig:shortdemo}
\end{figure}

\subsection{Main Results}
The perplexity performance of models on PG19 validation dataset is shown in Table ~\ref{tab:pg19}. As for the evaluation context length of 8192, LongQLoRA outperforms LongLoRA-LoRA and MPT-7B-8K, and is extremely close to LongLoRA-Full, only 0.03 higher. As for the perplexity of evaluation context length from 1024 to 4096 on PG19 validation dataset, LongQLoRA also slightly outperforms LongLoRA-LoRA and is close to LongLoRA-Full.

In Table ~\ref{tab:proof-pile}, we present the evaluation results on the Proof-pile test dataset. LongQLoRA also slightly outperforms LongLoRA-LoRA in evaluation context length from 1024 to 8192 and is also extremely close to LongLoRA-Full, almost the same as LongLoRA-Full in evaluation context length of 8192.

Note that MPT-7B-8K starts from MPT-7B, updates the sequence length to 8K and train for an additional 500B tokens, results in a total of 1.5T tokens of text and code, so it costs much more training GPUs with full finetuning. LongLoRA also costs 8 A100 GPUs though it has saved much training resources. LongQLoRA only use a single V100 GPU during finetuning which is more memory-efficient and can achieve close or even better performance.

LongQLoRA-Vicuna-13B-8K is finetuned based on Vicuna-13B whose context length is extended from 4096 to 8192. This model also achieves good performance in long context generartion and avoids the degration at short instruction following. We provide some examples generated by our model in Figure \ref{fig:longdemo} and Figure \ref{fig:shortdemo}.

\subsection{Ablation Study}
\paragraph{LoRA Rank.}
To study the effect of LoRA rank, we use different LoRA ranks to finetune LLaMA2-7B for 1000 steps. In Table ~\ref{tab:lora rank}, as the LoRA rank increases, the perplexity drop gradually. When LoRA rank is set to 64, LongQLoRA achieves almost the same performance as LongLoRA-Full and even slightly outperforms MPT-7B-8K, 0.02 lower. So 64 is a suitable setting of LoRA rank in LongQLoRA. Maybe as the LoRA rank continues to increase, the perplexity can further drop slightly.

\paragraph{Finetuning Steps.}
We also conduct an ablation experiment about the finetuning step on LLaMA2-7B, we evaluate the perplexity on PG19 validation dataset in evaluation context length of 8192 and the finetuning step from 0 to 1000. In Table ~\ref{tab:finetuning steps}, when the finetuning step is 0, that is to say, only extend context length with Position Interpolation and finetuning is not performed, the perplexity is high. Only after 100 finetuning steps, the perplexity drops dramatically. As the number of finetuning steps increases, the perplexity further decreases, and at about 1000 steps, the model basically converges and achieve good performance.

\begin{table}
	\caption{Ablation on attention patterns in inference. Models are finetuned based on LLaMA2-7B with shift short attention for 1000 steps with RedPajama dataset. The results are evaluated in perplexity on PG19 validation dataset. Standar global attention achieves better performance compared with shift short attention in inference.}
	\centering
	\begin{tabular}{ccc}
		\toprule
		Attention Pattern & PG19 & Proof-pile\\
		\midrule
		Shift Short Attention & 8.51  & 3.07    \\
		Standar Global Attention & 7.96 & 2.73  \\
		\bottomrule
	\end{tabular}
	\label{tab:attention pattern}
\end{table}
This ablation experiment further demonstrates the efficiency and effectiveness of LongQLoRA, which only requires finetuning with a small number of long text within 1000 steps to achieve good performance.

\paragraph{Attention Pattern in Inference.}
In Table ~\ref{tab:attention pattern}, we present the effect of different attention patterns on PG19 validation and Proof-pile test dataset during evaluation, including shift short attention and standar global attention.

As shown in Table ~\ref{tab:attention pattern}, perplexity is better when using standar global attention during inference, which indicates that shift short attention is not suitable in inference. It also shows that it is feasible to use shift short attention during finetuning and use standard attention during inference, it can be better compatible with existing inference framework.

\section{Conclusion}
In conclusion, we present LongQLoRA, an efficient and effective method to extend the context length of RoPE-based large language models. With LongQLoRA, we can extend context length of LLaMA2 7B and 13B to 8192 or 12k on a single V100 GPU with 32GB memory and achieve competitive performance compared with full finetuning. Because of the compatibility between shift short attention and standar global attention, models finetuned with LongQLoRA can be easily compatible with existing inference frameworks. We haven't further explore larger context length due to the limit of training resource, and we plan to investigate in the future.

\bibliographystyle{unsrtnat}
\bibliography{references}  






\end{document}